\documentclass{article}
\usepackage{spconf,amsmath,graphicx}
\usepackage{url}
\usepackage{enumitem}
\usepackage{multirow}
\usepackage[table,xcdraw]{xcolor}
\usepackage{booktabs}
\usepackage{adjustbox}
\usepackage{framed}
\usepackage{lscape}
\usepackage{multirow} 
\usepackage{subfiles}
\usepackage{hyperref}
\usepackage{makecell}
\usepackage[numbers,sort&compress,square]{natbib}

\title{Context and System Fusion in Post-ASR Emotion Recognition \\ with Large Language Models}
\name{Pavel Stepachev, Pinzhen Chen, Barry Haddow}
\address{School of Informatics, University of Edinburgh\\\texttt{\normalsize\{pavel.stepachev,pinzhen.chen,bhaddow\}@ed.ac.uk}}

\begin{document}
\maketitle
\begin{abstract}
Large language models (LLMs) have started to play a vital role in modelling speech and text. To explore the best use of context and multiple systems' outputs for post-ASR speech emotion prediction, we study LLM prompting on a recent task named GenSEC. Our techniques include ASR transcript ranking, variable conversation context, and system output fusion. We show that the conversation context has diminishing returns and the metric used to select the transcript for prediction is crucial. Finally, our best submission surpasses the provided baseline by 20\% in absolute accuracy.

\end{abstract}


\section{INTRODUCTION}
\label{sec:introduction}
With significant recent developments in large language models (LLMs), they appear to be a powerful tool for various tasks including speech and text problems \cite{gpt3_paper, pmlr-v202-radford23a}. Naturally, researchers have recently adopted them in speech emotion recognition (SER)  \cite{feng_foundation_2023}. The growing popularity of LLMs can be attributed to their capability as a general-purpose approach and their ready-to-use attribute. However, their performance remains modest in challenging conditions---e.g. the GPT-3.5 baseline at the GenSEC Task 3 merely sits at 55.18\% accuracy \cite{gensec}. While (re-)training an LLM for a certain task is prohibitive in many scenarios, we take this opportunity further to understand the optimal use of LLMs in this task by exploring LLM prompting in post-ASR SER.

GenSEC Task 3\footnote{Website: \href{https://sites.google.com/view/gensec-challenge/home}{sites.google.com/view/gensec-challenge/home}\\\phantom{000-}Task: \href{https://github.com/YuanGongND/llm_speech_emotion_challenge}{github.com/YuanGongND/llm\_speech\_emotion\_challenge}} provides various ASR systems' outcomes of conversations and participants are required to predict the speaker's emotion associated with certain utterances. From a machine learning perspective, it is straightforward to train a model especially to combine text and speech modalities \cite{sahu19_interspeech,Sebastian2019FusionTF,Pan2020MultimodalAF,li2022fusingasroutputsjoint,gong_lanser_2023,li2024crossmodal}. Yet, we put forward the motivation to \textit{prompt} a general-purpose LLM which is training-free: to prevent an algorithm from overfitting to unwanted data bias---e.g. choice of words by a speaker that is spuriously correlated with their emotion---which improves performance on a test set but not in a universal setting.

Formally, our approach explores suitable prompting strategies to perform speech emotion prediction from ASR outputs without speech signals. Most efforts are centred on creating a practical context for prompting. The contributions of this work are:
\begin{itemize}[topsep=0.1ex,itemsep=0.1ex]
\item Methodologically, we 1) select and rank ASR outputs as LLM input using multiple metrics and 2) exploit and fuse the conversation history and multiple ASR system outputs.
These approaches balance the trade-off between performance and cost of querying LLMs.
\item Performance-wise, we observe a huge leap from the provided baseline. Our final submission records an SER accuracy of 75.1\% surpassing the baseline by 20\%. Given our training-free paradigm, we expect it to be more generalizable to other settings.
\item To aid reproducibility, we make our code public.\footnote{Our implementation: \href{https://github.com/rggdmonk/GenSEC-Task-3}{github.com/rggdmonk/GenSEC-Task-3}}
\end{itemize}

\section{DATA AND EVALUATION}
\label{sec:dataset}

The task \cite{gensec} adopts IEMOCAP consisting of 5 sessions of scripted and improvised dialogues, all conducted in a controlled lab environment \cite{Busso2008IEMOCAPIE}. Four emotion classes are present: angry, happy (incl. excited), neutral, and sad. The training set has 5,225 utterances with 2,577 requiring predictions; the test set has 4,730 with 2,923 requiring predictions. The data includes outputs from 11 ASR systems \cite{li2024speech}: 
\begin{itemize} [noitemsep,topsep=0pt]
\item\texttt{hubertlarge}, 
\item\texttt{w2v2\{100$\vert$960$\vert$960large$\vert$960largeself\}}, \item\texttt{wavlmplus}, 
\item\texttt{whisper\{base$\vert$large$\vert$medium$\vert$small$\vert$tiny\}}.
\end{itemize}

In terms of constructing LLM prompting context, since the objective is to evaluate and improve SER performance in a realistic setting, we avoid using ground truth transcripts or emotion labels. We note that we use information from the dataset regarding each person's sex (female or male). Whilst many SER papers use 5 sessions for cross-validation, we follow a 10-session setup in this challenge's baseline configuration. The final system evaluation and team ranking adopts four-class unweighted accuracy.

\section{CONTEXT SELECTION METHODOLOGY}
\label{sec:best_cand_selection}
Conversation context and multiple ASR outputs are available in the GenSec dataset for emotion classification, so our investigation centres around forming a suitable context for prompting. Our first study is on picking a suitable ASR output for prompting, where we explore \textit{ranking} the ASR outputs using string-based metrics (Sec \ref{ssec:ranking}) and using handcrafted \textit{naive heuristics} for selection (Sec \ref{ssec:naive_heuristics}). The second direction is the use of conversation history represented by a variable size of context from ASR (Sec \ref{ssec: cw_and_last_pred}) as well as an examination of the feasibility of combining multiple ASR outputs (Sec \ref{ssec: cw_and_select}).

\subsection{Ranking}
\label{ssec:ranking}
A gold speech transcript is never available in practical scenarios. To select the most suitable transcript for emotion prediction, we perform ASR system output ranking based on inter-transcript aggregated metric scores, where each transcript is evaluated against all other outputs treated as references \cite{bogoychev-chen-2021-highs}. With $n$ transcripts and a metric function $\text{metric}()$, the aggregated score for each transcript $k$ is computed as:
\vspace{-1ex}
\begin{align}
\text{score}_{k} = \sum_{i = 0}^{n - 1} \text{metric}(\text{transcript}_{k}, \text{transcript}_{i}), i \ne k
\end{align}
\noindent All system outputs are then sorted by their aggregated scores. This can be viewed as simplified minimum Bayes risk decoding \cite{kumar-byrne-2004-minimum} with a uniform probability to pick a ``consensus'' ASR candidate. Concerning $\text{metric()}$ implementation, we explore an array of string measures: chrF and chrF++ from the sacreBLEU package\footnote{\href{https://github.com/mjpost/sacrebleu}{github.com/mjpost/sacrebleu} \tt v2.4.2} with default configurations \cite{post-2018-call,popovic-2015-chrf,popovic-2017-chrf}; word error rate (WER), match error rate (MER), word information lost (WIL), and word information preserved (WIP) \cite{Morris2004FromWA} from the JiWER package.\footnote{\href{https://github.com/jitsi/jiwer}{github.com/jitsi/jiwer} \tt v3.0.4}

\subsection{Naive selection heuristics}
\label{ssec:naive_heuristics}
We then propose simple heuristics for ASR candidate selection via various criteria:
\begin{itemize} [noitemsep,topsep=0pt]
\item\texttt{longest} selects the output with the highest character count, often containing the most content.
\item\texttt{shortest} selects the output with the lowest character count, having the least hallucination/verbosity.
\item\texttt{most\_punc} selects the output with the highest punctuation count, likely the most structured and expressive.
\item\texttt{least\_punc} selects the output with the least punctuation count, indicating simplicity/brevity.
\item\texttt{random} selects a random output for each utterance, giving a baseline with variety in ASR systems.
\end{itemize}

\noindent To enhance the robustness of these approaches, we also trial four composite metrics by matching (\texttt{longest}, \texttt{shortest}) with (\texttt{most\_punc}, \texttt{least\_punc}) where length is the primary constraint. Technically, spaces are ignored during character counting and the punctuation list is pre-defined as \texttt{!?.,;:-\$\%\&}.

\subsection{The use of conversation context}
\label{ssec: cw_and_last_pred}

Moving on from ASR output to previous conversation history, we conduct a set of experiments with varying context window ({CW}) sizes, to assess its impact on emotion prediction accuracy. Longer context provides information on the dialogue, yet the texts come from ASR systems instead of ground truths, bringing in accumulated noise. In addition, we designed an LLM prompt template to simulate a natural real-life scenario as shown in Figure~\ref{fig:prompt_cw}.

\begin{figure}[t]
\footnotesize
\begin{framed}
\textit{Below is a transcript of a conversation between a male and a female:}
\newline \textit{Person A (female):} \{selected\_candidate\}
\newline \textit{Person B (male):} \{selected\_candidate\}
\newline \textit{Person A (female):} \{selected\_candidate\}
\newline
\newline \textit{I need help understanding the emotional context of the last line. As a non-native English speaker, this is very important to me. Could you identify the emotion expressed in the last utterance (anger, happiness, sadness, or neutral)? Please provide a brief explanation for your choice. Select a single emotion and enclose it in square brackets, like this: [emotion]. The emotion can only be anger, happiness, sadness, or neutral.}
\end{framed}
    \caption{Prompt template with context size 2 with the last utterance needing emotion prediction.}
    \label{fig:prompt_cw}
\end{figure}

\subsection{ASR output fusion}
\label{ssec: cw_and_select}
Finally, we examine the feasibility of using multiple ASR outputs simultaneously since different systems could capture different nuances in the original speech and also make different error patterns that might be used to alleviate each other. For instance, we observe that \texttt{whisper-large} tends to generate longer ``hallucinations'' than \texttt{whisper-tiny} potentially due to an overly strong internal language model. In our experiments, we fix the rough input length to LLM calls by varying a combination of (CW, N)---where CW is the context size, N is the number of ASR outputs, and their sum remains constant. To facilitate the inclusion of multiple ASR outputs, the study uses an updated prompt in Figure~\ref{fig:prompt_cw_n}.

\begin{figure}[h!t]
\footnotesize
\begin{framed}
\textit{Below is a transcript of a conversation between a male and a female:}
\newline \textit{Person B (male):} \{selected\_candidate\}
\newline \textit{Person A (female):} \{selected\_candidate\}
\newline \textit{Person B (male):} \{selected\_candidate\}
\newline \textit{Person A (female):} \{selected\_candidate\}
\newline
\newline \textit{I am not a native English speaker and I did not hear the last utterance from Person B (male) very clearly. It could be one of the following:}
\newline\{random\_unique\_asr\_output\}\textit{, or}
\newline\{random\_unique\_asr\_output\}\textit{, or}
\newline\{selected\_candidate\}\textit{, or}
\newline\{random\_unique\_asr\_output\}\textit{, or}
\newline\{random\_unique\_asr\_output\}\textit{.}
\newline
\newline \textit{It is now very important for me to understand the emotion of Person B (male) from your choice. Could you identify the emotion expressed in the last utterance (anger, happiness, sadness, or neutral)? Please provide a brief explanation for your choice. Select a single emotion and enclose it in square brackets, like this: [emotion]. The emotion can only be anger, happiness, sadness, or neutral.}
\end{framed}
    \caption{Prompt template with a context size 4 as well as 5 ASR outputs as a means of fusion.}
    \label{fig:prompt_cw_n}
\end{figure}

\section{EXPERIMENTS AND RESULTS}
\label{sec:exp_setup}
\subsection{Setup}
In our exploration of LLM prompting for speech emotion recognition, we use OpenAI models through the API.
We tried two models \texttt{gpt-4o} and \texttt{gpt-3.5-turbo}.\footnote{Specifically, gpt-4o-2024-05-13 and gpt-3.5-turbo-0125 with temperature=1, max\_tokens=250, seed=42.} The system message remained unchanged: ``You are a helpful assistant.'' All reported experimental results are on the training set; final test predictions are scored by the task organizers.

\subsection{Results}
\label{ssec:results}

\subsubsection{Comparison between GPT models}
\label{sssec:Comparison between GPT models}
We performed an initial experiment on \texttt{gpt-3.5-turbo} and \texttt{gpt-4o} with varying context sizes. Table~\ref{tab:model_comp} summarizes the results showing that \texttt{gpt-4o} consistently outperforms \texttt{gpt-3.5-turbo} in accuracy across all scenarios. Our further experiments all use \texttt{gpt-4o}.

\begin{table}[h!t]
\centering\small
\begin{tabular}{@{}ccc@{}}
\toprule
\textbf{CW} & \texttt{gpt-3.5-turbo} & \texttt{gpt-4o} \\ \midrule
0           & 0.386         & 0.449  \\
2           & 0.414         & 0.510   \\
4           & 0.437         & 0.545  \\
8           & 0.439         & 0.555  \\
16          & 0.472         & 0.575  \\
32          & 0.475         & 0.574  \\ \bottomrule
\end{tabular}
\caption{Comparison between \texttt{gpt-3.5-turbo} and \texttt{gpt-4o} at various context sizes. The selection metric is \texttt{longest\_and\_most\_punc} and the evaluation metric is {unweighted accuracy}.}
\label{tab:model_comp}
\end{table}

\subsubsection{Effects of the conversation context}
\label{sssec:Conversation context size}

Referring back to Table~\ref{tab:model_comp}, both models demonstrate increasing accuracy as the context expands, with the most significant improvements occurring at smaller window sizes (0 to 4). However, improvement diminishes as the context window size continues to increase.

\begin{figure}[t]
    \centering\small
    \includegraphics[width=\columnwidth]{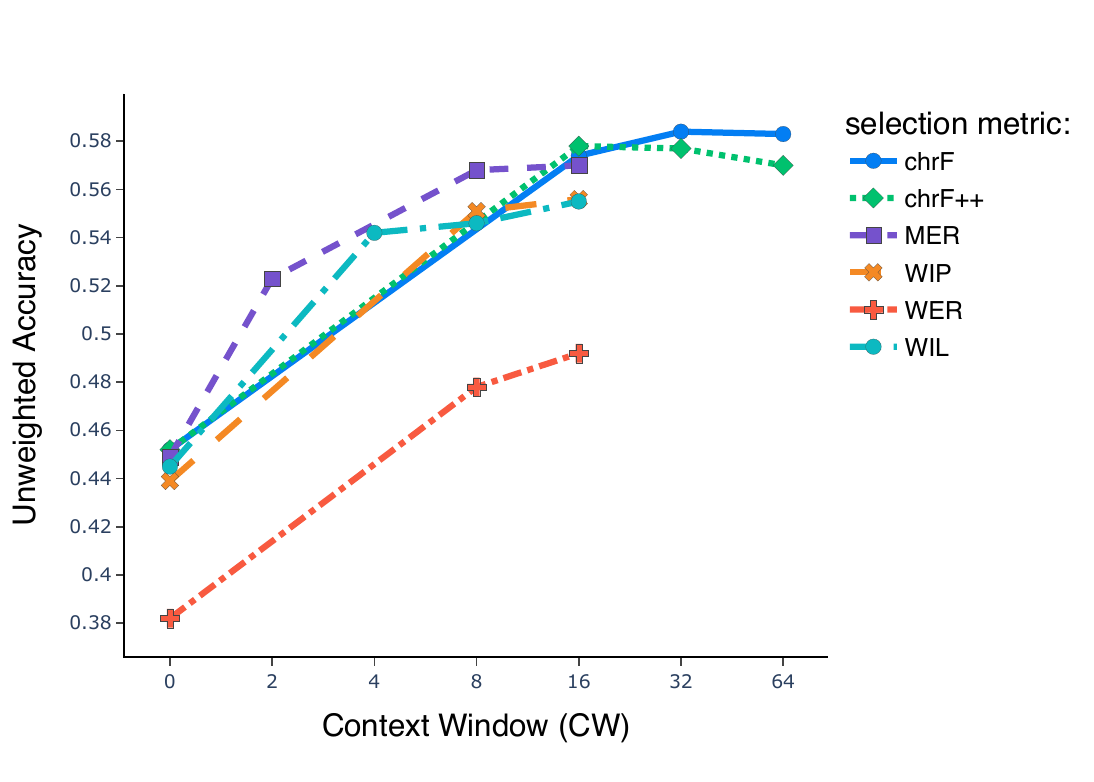}
    \caption{Performance of {ranking} metrics with various context sizes on \texttt{gpt-4o}.}\vspace{-1ex}
    \label{fig:acc_vs_cw_ranking}
\end{figure}

Figure~\ref{fig:acc_vs_cw_ranking} illustrates the impact of \textit{ranking} metrics on unweighted accuracy as the context size increases. Among these, chrF and chrF++ consistently show the highest numbers, particularly as the context window size grows beyond 8. MER follows closely behind but slightly underperforms compared to chrF and chrF++. On the other hand, WER consistently lags, showing a significant gap in accuracy, especially at larger CW sizes. WIP and WIL show moderate performance, with WIP slightly outperforming WIL.

\begin{figure}[t]
    \centering\small
    \includegraphics[width=\columnwidth]{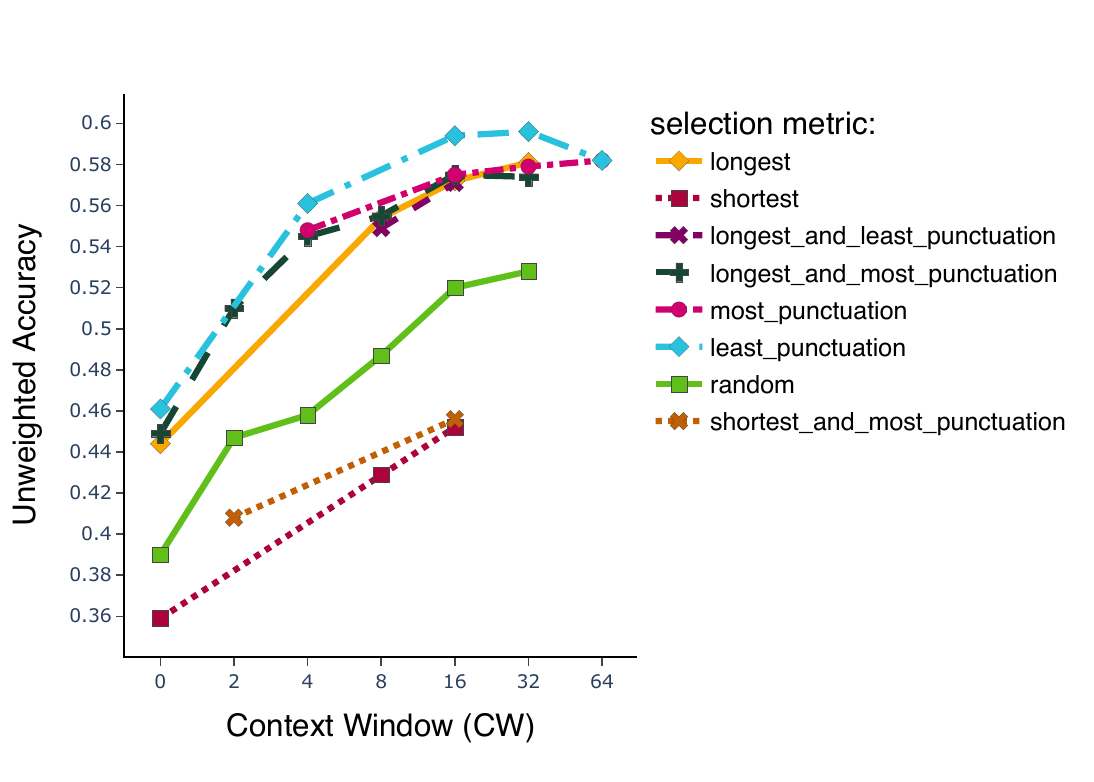}
    \caption{Performance of naive heuristics metrics with various context sizes on \texttt{gpt-4o}.}\vspace{-1ex}
    \label{fig:acc_vs_cw_nh}
\end{figure}

Figure~\ref{fig:acc_vs_cw_nh} shows the same for \textit{naive heuristics}. The metrics associated with the longest text and punctuation generally yield the highest accuracy, with significant improvements as the CW size increases. Specifically, the \texttt{longest\_and\_least\_punc}, \texttt{longest\_and\_most\_punc}, \texttt{longest}, and \texttt{most\_punc} metrics cluster at the top, indicating superior performance. The \texttt{least\_punc} metric starts lower but nearly catches up at larger CW sizes. In contrast, the \texttt{random} selection shows moderate improvement but remains below the top-performing metrics. The \texttt{shortest} and \texttt{shortest\_and\_most\_punc} metrics exhibit the lowest performance, with a relatively flat increase in accuracy, suggesting a limited benefit from increasing CW size.

\begin{table}[h!]
\centering\small\setlength{\tabcolsep}{5ex}
\begin{tabular}{@{}lc@{}}
\toprule
\multicolumn{1}{c}{\textbf{Metric}}                           & \textbf{Acc}      \\ \midrule
chrF                             & \textbf{0.574} \\
chrF++                & \textbf{0.578} \\
WER                              & 0.492 \\
MER                              & \textbf{0.570}  \\
WIL                              & 0.555 \\
WIP                              & 0.556 \\ \midrule
\texttt{random}                           & 0.520  \\
\texttt{longest}                          & 0.572 \\
\texttt{shortest}                         & 0.452 \\
\texttt{longest\_and\_least\_punc} & 0.571 \\
\texttt{longest\_and\_most\_punc}  & \textbf{0.575} \\
\texttt{most\_punc}                & \textbf{0.575} \\
\texttt{least\_punc}               & \textbf{0.594} \\
\texttt{shortest\_and\_most\_punc} & 0.456  \\ \bottomrule
\end{tabular}
\caption{Comparison between {ranking} and {naive heuristics} in ASR candidate selection with a context window of $16$ and \texttt{gpt-4o}.}
\label{tab:only_cw_16}
\end{table}

In both figures, increasing the context window generally leads to higher accuracy, though the degree of improvement varies depending on the selection metric. Metrics that prioritize longer text segments and punctuation factors tend to perform better, underscoring the importance of these elements in achieving higher accuracy in this challenge.

\subsubsection{ASR output ranking}
\label{sssec:ASR output ranking}
Table~\ref{tab:only_cw_16} compares selection metrics using a fixed context window and the same model. Out of the ranking metrics, chrF++ stands out as one of the top performers, demonstrating high accuracy. The chrF, MER, WIL, and WIP metrics also show relatively high accuracy, although slightly lower than chrF++. In contrast, WER shows notably poor performance with lower accuracy. For naive heuristics, the \texttt{least\_punc} method achieves the highest overall accuracy. 
The \texttt{most\_punc} and \texttt{longest\_and\_most\_punc} heuristics also perform well, with each showing high accuracy. On the other hand, the \texttt{shortest\_and\_most\_punc} and \texttt{shortest} metrics yield the lowest accuracy, indicating poorer performance. In summary, punctuation-related metrics tend to deliver better accuracy, with \texttt{least\_punc} emerging as the leading method.

\subsubsection{ASR-context fusion}
\label{sssec: ASR-context fusion}
To explore the best use of the prompting context, we evaluate various combinations of context window sizes and the ASR candidates. The results, as detailed in the Table \ref{tab:cn_all_runs}, show that the highest unweighted accuracy for the \texttt{longest\_and\_most\_punc} strategy is achieved with a CW of 29 and 3 candidates. The highest accuracy is obtained with a CW of 27 and 5 candidates for the \texttt{least\_punc} strategy. These findings suggest that the model performs better with a larger context window and more candidates, particularly when the context has minimal punctuation.

\begin{table}[t]
\centering\small\setlength{\tabcolsep}{2ex}
\begin{tabular}{@{}cccc@{}}
\toprule
 \multicolumn{1}{l}{\textbf{CW}} & \multicolumn{1}{l}{\textbf{N}} & \multicolumn{1}{c}{\makecell{\texttt{longest\_and\_}\\\texttt{most\_punc}}} & \multicolumn{1}{c}{\texttt{least\_punc}}  \\ \midrule
2                      & 1                     & 0.514                                            & \textbf{0.529} \\
1                      & 2                     & 0.486                                            & 0.492                                  \\ \cmidrule(lr){1-4}
4                      & 1                     & 0.543                                            & \textbf{0.558} \\
2                      & 3                     & 0.506                                            & 0.513                                  \\ \cmidrule(lr){1-4}
8                      & 1                     & 0.560                                             & \textbf{0.583} \\
6                      & 3                     & 0.565                                            & 0.570                                   \\
4                      & 5                     & 0.546                                            & 0.564                                  \\ \cmidrule(lr){1-4}
29                     & 3                     & 0.594                                            & 0.596                                  \\
27                     & 5                     & 0.581                                            & \textbf{0.606} \\
\bottomrule
\end{tabular}%
\caption{Performance of context-ASR system fusion with \texttt{gpt-4o}. \textbf{CW} denotes the size of the context window and \textbf{N} denotes the number of ASR candidates.}
\label{tab:cn_all_runs}
\end{table}

\begin{table}[t]
\centering\small
\begin{tabular}{lrlr}
\toprule
 & \multicolumn{1}{c}{Acc} & \multicolumn{1}{c}{Configuration}                            &  \multicolumn{1}{c}{Session} \\
 \midrule
baseline     & 0.551                   & CW=3 $\vert$ \texttt{whispertiny} & 10 \\
ours 1 & \textbf{0.751}          & CW=29 \& N=3 $\vert$ \texttt{least\_punc}                & 10                     \\
ours 2 & 0.744                   & CW=27 \& N=5 $\vert$ \texttt{least\_punc}              & 10                     \\
ours 3 & 0.742                   & CW=64 $\vert$ \texttt{most\_punc}  & 5                      \\ 
ours 4 & 0.725                   & CW=16 $\vert$ \texttt{least\_punc} & 5                      \\
ours 5 & 0.724                   & CW=32 $\vert$ \texttt{chrF}               & 5                      \\
\bottomrule
\end{tabular}%
\caption{The official results on test set. Baseline uses \texttt{gpt-3.5-turbo}; our submissions use \texttt{gpt-4o}.}
\label{tab:sub_results}
\end{table}

\subsubsection{Final results}
\label{sssec: Shared task submissions}

Based on findings from the training data, we chose five configurations to run on the test split. These are then evaluated by the authors of GenSec \cite{gensec}; we have no access to test set references. Table \ref{tab:sub_results} outlines our official results---they all surpass the provided baseline by a large margin of 20\% absolute accuracy. Finally, a comprehensive breakdown of all experiments can be found in Appendix~\ref{app:appendix} Table~\ref{tab:cw_all_runs}.

\section{CONCLUSION}
\label{sec:conclusion}
In this paper, we explore LLM prompting for speech emotion prediction which is training-free. Our methods include ASR candidate selection and context fusion that does not rely on audio signals or ground truth transcripts. Evaluations on the IEMOCAP dataset, as re-split by the organizers, demonstrate that our approach achieves a strong result of 75.1\%, outperforming the baseline by 20\% accuracy. Notably, our methods, being training-free, mitigate the risk of overfitting to speaker-specific or ASR system-specific biases.

\section{ACKNOWLEDGMENTS}
\label{sec:ack}
We thank the GenSec authors for their assistance. This work was funded by UK Research and Innovation (UKRI) under the UK government’s Horizon Europe funding guarantee [grant numbers 10052546 and 10039436]. 
The authors would like to thank EDINA and the Information Services Group at the University of Edinburgh for generously providing OpenAI credits that supported this research.

\bibliographystyle{IEEEbib}
\bibliography{refs}

\appendix


\begin{table*}[htbp]
\section{APPENDIX}

\label{app:appendix}
\centering
\footnotesize
\begin{adjustbox}{width=0.6\textwidth}
\footnotesize
\begin{tabular}{@{}clllll@{}}
\toprule
\multicolumn{1}{c}{\textbf{Model}}                          & \multicolumn{1}{c}{\textbf{Selection Metric}}                         & \textbf{CW} & \textbf{Acc}      & \textbf{CW Ranking} & \textbf{Ranking} \\ \midrule
\multirow{54}{*}{\texttt{gpt-4o}} & \multirow{4}{*}{chrF}                             & 0  & 0.452±0.019 & 2-3 & 43-45 \\
                         &                                                   & 16 & 0.574±0.019 & 5   & 13-14 \\
                         &                                                   & 32 & 0.584±0.019 & 2   & 3     \\
                         &                                                   & 64 & 0.583±0.019 & 1   & 4     \\ \cmidrule(l){2-6} 
                         & \multirow{4}{*}{chrF++}                 & 0  & 0.452±0.019 & 2-3 & 43-45 \\
                         &                                                   & 16 & 0.578±0.019 & 2   & 9     \\
                         &                                                   & 32 & 0.577±0.019 & 5   & 10    \\
                         &                                                   & 64 & 0.57±0.019  & 4   & 17-18 \\ \cmidrule(l){2-6} 
                         & \multirow{5}{*}{\texttt{least\_punc}}               & 0  & 0.461±0.019 & 1   & 40    \\
                         &                                                   & 4  & 0.561±0.019 & 1   & 20    \\
                         &                                                   & 16 & 0.594±0.019 & 1   & 2     \\
                         &                                                   & 32 & 0.596±0.019 & 1   & 1     \\
                         &                                                   & 64 & 0.582±0.019 & 2-3 & 5-6   \\ \cmidrule(l){2-6} 
                         & \multirow{4}{*}{\texttt{longest}}                          & 0  & 0.444±0.019 & 7   & 50    \\
                         &                                                   & 8  & 0.554±0.019 & 3   & 24    \\
                         &                                                   & 16 & 0.572±0.019 & 6   & 15    \\
                         &                                                   & 32 & 0.581±0.019 & 3   & 7     \\ \cmidrule(l){2-6} 
                         & \multirow{2}{*}{\texttt{longest\_and\_least\_punc}} & 8  & 0.549±0.019 & 5   & 26    \\
                         &                                                   & 16 & 0.571±0.019 & 7   & 16    \\ \cmidrule(l){2-6} 
                                                   & \multirow{6}{*}{\texttt{longest\_and\_most\_punc}} & 0  & 0.449±0.019 & 4-5     & 46-47         \\
                         &                                                   & 2  & 0.51±0.019  & 2   & 34    \\
                         &                                                   & 4  & 0.545±0.019 & 3   & 29    \\
                         &                                                   & 8  & 0.555±0.019 & 2   & 22-23 \\
                         &                                                   & 16 & 0.575±0.019 & 3-4 & 11-12 \\
                         &                                                   & 32 & 0.574±0.019 & 6   & 13-14 \\ \cmidrule(l){2-6} 
                         & \multirow{4}{*}{MER}                              & 0  & 0.449±0.019 & 4-5 & 46-47 \\
                         &                                                   & 2  & 0.523±0.019 & 1   & 32    \\
                         &                                                   & 8  & 0.568±0.019 & 1   & 19    \\
                         &                                                   & 16 & 0.57±0.019  & 8   & 17-18 \\ \cmidrule(l){2-6} 
                         & \multirow{4}{*}{\texttt{most\_punc}}                & 4  & 0.548±0.019 & 2   & 27    \\
                         &                                                   & 16 & 0.575±0.019 & 3-4 & 11-12 \\
                         &                                                   & 32 & 0.579±0.019 & 4   & 8     \\
                         &                                                   & 64 & 0.582±0.019 & 2-3 & 5-6   \\ \cmidrule(l){2-6} 
                         & \multirow{6}{*}{\texttt{random}}                           & 0  & 0.39±0.019  & 9   & 57    \\
                         &                                                   & 2  & 0.447±0.019 & 3   & 48    \\
                         &                                                   & 4  & 0.458±0.019 & 5   & 41    \\
                         &                                                   & 8  & 0.487±0.019 & 7   & 36    \\
                         &                                                   & 16 & 0.52±0.019  & 11  & 33    \\
                         &                                                   & 32 & 0.528±0.019 & 7   & 31    \\ \cmidrule(l){2-6} 
                         & \multirow{3}{*}{\texttt{shortest}}                         & 0  & 0.359±0.018 & 12  & 60    \\
                         &                                                   & 8  & 0.429±0.019 & 10  & 54    \\
                         &                                                   & 16 & 0.452±0.019 & 15  & 43-45 \\ \cmidrule(l){2-6} 
                         & \multirow{2}{*}{\texttt{shortest\_and\_most\_punc}} & 2  & 0.408±0.019 & 5   & 56    \\
                         &                                                   & 16 & 0.456±0.02  & 14  & 42    \\ \cmidrule(l){2-6} 
                         & \multirow{3}{*}{WER}                              & 0  & 0.382±0.019 & 11  & 59    \\
                         &                                                   & 8  & 0.478±0.02  & 8   & 37    \\
                         &                                                   & 16 & 0.492±0.019 & 12  & 35    \\ \cmidrule(l){2-6} 
                         & \multirow{4}{*}{WIL}                              & 0  & 0.445±0.019 & 6   & 49    \\
                         &                                                   & 4  & 0.542±0.02  & 4   & 30    \\
                         &                                                   & 8  & 0.546±0.02  & 6   & 28    \\
                         &                                                   & 16 & 0.555±0.019 & 10  & 22-23 \\ \cmidrule(l){2-6} 
                         & \multirow{3}{*}{WIP}                              & 0  & 0.439±0.019 & 8   & 51-52 \\
                         &                                                   & 8  & 0.551±0.019 & 4   & 25    \\
                         &                                                   & 16 & 0.556±0.019 & 9   & 21    \\ \midrule
\multicolumn{1}{l}{\multirow{7}{*}{\texttt{gpt-3.5-turbo}}} & \multirow{6}{*}{\texttt{longest\_and\_most\_punc}} & 0  & 0.386±0.019 & 10      & 58            \\
\multicolumn{1}{l}{}     &                                                   & 2  & 0.414±0.019 & 4   & 55    \\
\multicolumn{1}{l}{}     &                                                   & 4  & 0.437±0.019 & 6   & 53    \\
\multicolumn{1}{l}{}     &                                                   & 8  & 0.439±0.019 & 9   & 51-52 \\
\multicolumn{1}{l}{}     &                                                   & 16 & 0.472±0.019 & 13  & 39    \\
\multicolumn{1}{l}{}     &                                                   & 32 & 0.475±0.02  & 8   & 38    \\ 
     &    baseline \cite{gensec}                                                & 3 & 0.447\phantom{±0.000}  &    &    \\ \bottomrule
\end{tabular}
\end{adjustbox}
\caption{A list of all results on the training set. The evaluation metric is \texttt{unweighted accuracy} with two-sided bias-corrected and accelerated (BCa) bootstrap confidence intervals. \textbf{Ranking} denotes the global ranking of the configuration; \textbf{CW ranking} denotes rank within the same context size. } 
\label{tab:cw_all_runs}
\end{table*}

\end{document}